\documentclass[runningheads,a4paper]{llncs}

\usepackage{amssymb}
\usepackage{graphicx}

\usepackage{url}

\usepackage{wrapfig}
\usepackage{amsmath}
\usepackage{booktabs}
\usepackage{subcaption} % For subfigure
\usepackage{color}
\usepackage{hyperref}
\usepackage{multirow}

\usepackage{dsfont} % KK: For \mathds{1} indicator function
\newcommand{\Reals}{\mathds{R}}

\usepackage{diagbox} % for the diagonal in the table.

\begin{document}

\mainmatter  % start of an individual contribution

%\title{Learning from heterogeneously labeled data with CNNs for abdominal organ segmentation}
%\title{Learning under Annotation Shift:\\Exploiting Heterogeneous Data in Segmentation}

%\title{Learning from Partially Overlapping and Contradicting Labels in Image Segmentation}

\title{Learning from Partially Overlapping Labels: Image Segmentation under Annotation Shift}

\titlerunning{Learning from Partially Overlapping Labels}


\author{Gregory Filbrandt\inst{1}$^*$  \and
Konstantinos Kamnitsas\inst{1}$^*$ \and
David Bernstein \inst{2} \and \\
Alexandra Taylor \inst{3}  \and
Ben Glocker \inst{1}}
\authorrunning{Filbrandt, Kamnitsas, et al.}

% Submissions are initially anonymous. Authors and affiliations added only at camera ready after acceptance.
\institute{Department of Computing, Imperial College London, UK \and
Joint Dept of Physics, The Institute of Cancer Research and The Royal Marsden NHS Foundation Trust, UK \and
Department of Radiotherapy, The Royal Marsden NHS Foundation Trust, UK \\
$^*$ Equal contribution
}

%\toctitle{Lecture Notes in Computer Science}
%\tocauthor{Authors' Instructions}
\maketitle

% =========================================================================
% =========================================================================

% 8 pages of content (includingtext, figures, and tables) plus up-to 2 pages of references

\begin{abstract}

Scarcity of high quality annotated images remains a limiting factor for training accurate image segmentation models. While more and more annotated datasets become publicly available, the number of samples in each individual database is often small. Combining different databases to create larger amounts of training data is appealing yet challenging due to the heterogeneity as a result of differences in data acquisition and annotation processes, often yielding incompatible or even conflicting information. In this paper, we investigate and propose several strategies for learning from partially overlapping labels in the context of abdominal organ segmentation. We find that combining a semi-supervised approach with an adaptive cross entropy loss can successfully exploit heterogeneously annotated data and substantially improve segmentation accuracy compared to baseline and alternative approaches.

\end{abstract}

% =========================================================================

% =========================================================================
\section{Introduction}
\label{sec:intro}

\begin{figure}[t]
% \captionsetup{font=scriptsize}
\centering
\begin{subfigure}[b]{0.5\textwidth}
	\centering
	\includegraphics[clip=true, trim=0pt 0pt 0pt 0pt, width=1.0\textwidth]{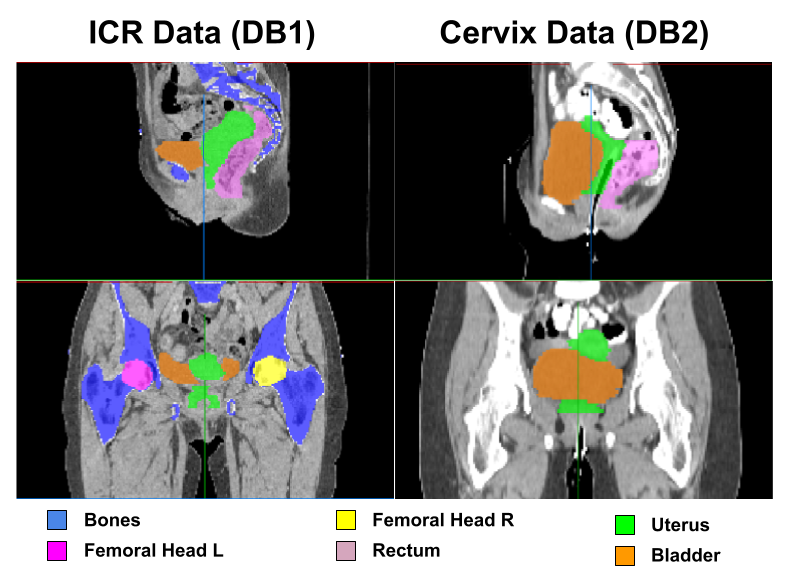}
	\caption{Heterogeneous labels in abdominal CT}
	\label{fig:intro1}
\end{subfigure}
\begin{subfigure}[b]{0.475\textwidth}
	\centering
	\includegraphics[clip=true, trim=0pt 0pt 0pt 0pt, width=1.052\textwidth]{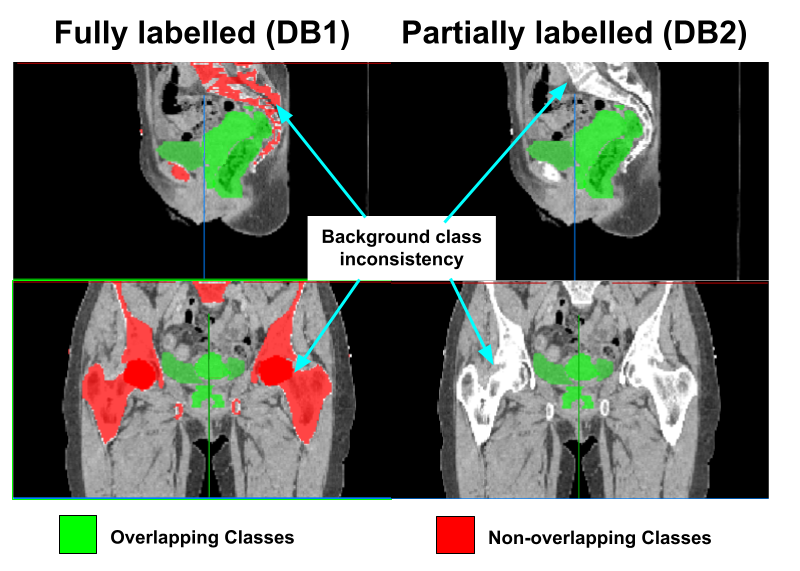}
	\caption{Label-contradiction problem}
	\label{fig:intro2}
\end{subfigure}
\caption{a) Annotation shift between the two databases in this study. DB1 has $6$ structures annotated, whereas only $3$ of them are annotated on DB2. b) Label contradiction problem due to a structure being labeled differently in the two databases (e.g. spine is part of background in DB2).}
\label{fig:intro}
\end{figure}

Obtaining sufficient amounts of high quality and accurate annotations in the context of image segmentation remains a major bottleneck due to the time-consuming nature of the expert labelling task. In recent years, an increasing amount of publicly available data has become available (e.g., brain MRI \cite{marcus2007open} or abdominal CT \cite{cervixDatabase}), often through the efforts of organizing computational challenges and benchmarks. However, these public datasets are often either limited in size or specific to a particular anatomy or pathology of interest (e.g., brain \cite{menze2014multimodal} or liver tumours \cite{bilic2019liver}). Pooling data from different studies to form larger datasets that are suitable for training automated segmentation methods is appealing yet challenging, due to the inherent heterogeneity of the available annotations. The set of labels from different datasets may be partially overlapping, but more importantly, may yield conflicting information due to differences in the annotation protocols (a problem also known as `annotation shift' \cite{castro2020causality}). Learning from such heterogeneous data is an open problem in machine learning for imaging.

Here, we investigate this challenge of learning under annotation shift in the context of automated segmentation of abdominal CT for the application of radiotherapy planning. In this study, we consider two datasets, one internal and one external, with partially overlapping and contradicting labels, as illustrated in Fig.~\ref{fig:intro}. Our goal is to devise learning strategies that can successfully exploit the available information from the different datasets with the aim to improve segmentation accuracy. While focusing on abdominal CT, our results should be of interest for other medical imaging applications and modalities.

\textbf{Related work:} Learning from heterogeneous data poses a variety of challenges. Previous works investigated learning in the existence of different input distributions across databases, known as the domain (or acquisition) shift. This was approached via domain adaptation \cite{kamnitsas2017unsupervised}, augmentation \cite{dorent2021learning}, feature-matching \cite{dou2020unpaired} or their combination.
This study instead focuses on the issue of annotation shift, where we assume that the total set of labels is $\mathcal{Y}$, but each database $\mathcal{D}_k$ has been annotated with a possibly different subset of labels, $\mathcal{Y}_k \subseteq \mathcal{Y}$. This problem has been commonly explored under the assumption that the label sets $\mathcal{Y}_k$ are \emph{disjoint} \cite{dorent2021learning,roulet2019joint}, which can be formulated as \emph{multi-task learning} \cite{moeskops2016deep,rajchl2018neuronet,yan2020learning,dorent2021learning}. A common approach is to construct a model that predicts a separate output for each label set $\mathcal{Y}_k$, for example, using a multi-head neural net. If a single output is desired (eg. a joint segmentation map) then the multiple outputs need to be fused via task-specific choice of aggregation rules (e.g. brain lesion taking precedence over a brain anatomy segmentation map \cite{dorent2021learning}). The multi-task approach has also been tried for problems where label sets $\mathcal{Y}_k$ may partially overlap. In this case, a means for fusing the separate output per label set in a single prediction is required, often done as a post-processing step, such as Non Maximal Suppression \cite{yan2020learning}. These fusion steps that are external to the model do not facilitate learning.
Instead of altering the model to predict a different output per label set, an alternative approach is to predict outputs that can take any value of the total label set, $\mathcal{Y}$, and develop a learning method that can process data with annotations belonging to different subsets $\mathcal{Y}_k$. Such a training objective for learning with two disjoint label sets has been proposed in \cite{roulet2019joint}, termed \emph{adaptive cross entropy} (ACE). When making predictions for training image $x_i^k$ with manual annotation $y_i^k$ from label set $\mathcal{Y}_k$, it considers the labels outside its own label set ($\mathcal{Y}_{o,k}\!=\!\mathcal{Y} - \mathcal{Y}_k$) as part of the background class, alleviating the label contradiction in the application of the cost function (Fig.~\ref{fig:intro}). This method has been originally proposed under the assumption that the label sets are disjoint, specifically for joint learning of brain structures and lesions from different databases. The original ACE, however, does not facilitate improvements for non-overlapping labels $\mathcal{Y}_{o,k}$ when processing background pixels, which we build upon.

\textbf{Contributions:} 
%This study explores learning a segmentation model that predicts belonging to total label set $\mathcal{Y}$, and label-sets $\mathcal{Y}_k$ from databases $\mathcal{D}_k$ may be partially overlapping ($\mathcal{Y}_k \!\cap\! \mathcal{Y}_l \! \neq \!\varnothing$). 
This study explores how to train a segmentation model to predict output $y\!\in\!\mathcal{Y}$ from the total label-set, using databases $\mathcal{D}_k$ with label-sets $\mathcal{Y}_k$ that may partially overlap, $\mathcal{Y}_k \!\cap\! \mathcal{Y}_l \! \neq \!\varnothing$.
We adapt ACE to this setting and extend it by interpreting voxels of the background class as unlabeled samples, adopting ideas from semi-supervised learning to improve learning from them. We find that ACE facilitates learning of overlapping classes, whereas semi-supervision provides benefits by also regularizing non-overlapping classes. Experiments with two databases of abdominal CT with partially overlapping label sets show that combining these complementary approaches improves segmentation.

% ====================================================================
\section{Learning from heterogeneously labeled data}
\label{sec:main_methods}

\begin{figure}[t]
% \captionsetup{font=scriptsize}
\centering
\begin{subfigure}[b]{0.31\textwidth}
	\centering
	\includegraphics[clip=true, trim=0pt 0pt 0pt 0pt, width=1.0\textwidth]{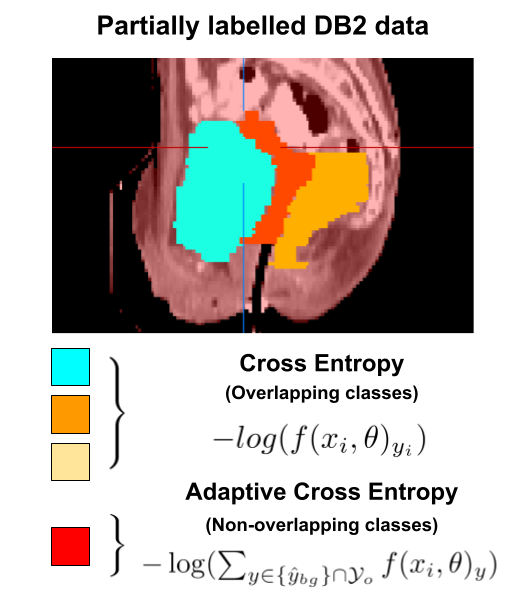}
	\caption{ACE}
	\label{fig:ace}
\end{subfigure}
\begin{subfigure}[b]{0.68\textwidth}
	\centering
	\includegraphics[clip=true, trim=0pt 0pt 0pt 0pt, width=1.0\textwidth]{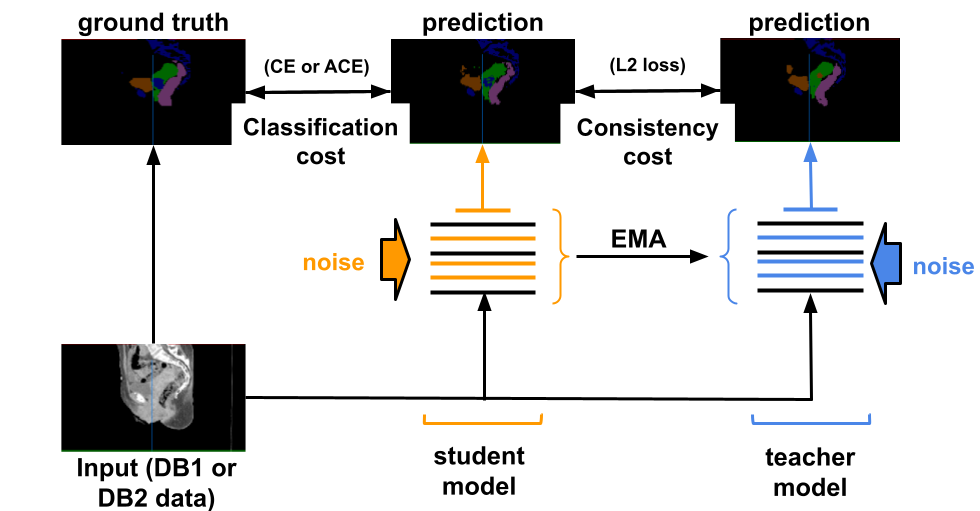}
	\caption{Mean teacher}
	\label{fig:mt}
\end{subfigure}
\caption{a) Adaptive Cross Entropy (ACE) loss as it varies between overlapping and non-overlapping classes. b) Key components of the Mean Teacher \cite{tarvainen2017mean}, one of the SSL approaches explored in this study.
%In training, each input sample is passed through a student and teacher model applying differing augmentations to the data. The output predictions are compared using a consistency cost through the calculation of the L2 loss. Additionally to encourage correct learning the prediction of the student model, for DB1 data, is evaluated against the ground truth labels using a classification cost. The classification cost, dependent on the implementation, may be either the cross entropy loss (CEL) or the adaptive cross entropy loss (ACE). In the case of the CEL this is only applied to fully labeled DB1 data. For the ACE loss both fully labeled DB1 and partially labeled DB2 data may be used. These two losses are evaluated for each training step for all the training data.
}
\label{fig:methods}
\end{figure}

%We first formalize the problem and describe the label-contradiction arising when different sets of labels are annnotated per database (\ref{subsec:baselines}). We then describe adaptive cross entropy (ACE) \cite{roulet2019joint}, originally proposed for disjoint labelsets, which we show it is applicable to the general case for arbitrarily overlapping label-sets with a subtle, more general re-formulation (Sec. \ref{subsec:ace}). We then describe semi-supervised learning (SSL) methods such as pseudo-labelling  (\ref{subsec:pseudolabels}) and mean-teacher Sec.\ref{subsec:mean_teacher}, which we show offer orthogonal benefits to ACE.

\subsection{Problem definition and label-contradiction issue}
\label{subsec:baselines}

Assume a set of classes of interest described by the \emph{total} label-set $\mathcal{Y}$, with cardinality $|\mathcal{Y}|\!=\!C$ number of classes. We define as $\mathcal{D}_k=\{(x_i^k,y_i^k)\}_{i=1}^{N_k}$ different databases, where $x_i^k$ is the $i$-th sample (image for classification or pixel for segmentation) of $\mathcal{D}_k$ and $y_i^k$ its true label. We do not assume disjoint label-sets as in previous works \cite{roulet2019joint,dorent2021learning}, but rather investigate the general case where only a subset $\mathcal{Y}_k \subseteq \mathcal{Y}$ of labels has been annotated for the $k$-th database, and different label-sets $\mathcal{Y}_k$ may be overlapping arbitrarily. We denote with $\mathcal{Y}_{o,k}=\mathcal{Y} - \mathcal{Y}_k$ the set of labels that have \emph{not} been annotated in database $k$. Finally, we define as ${\hat{y}_{bg}}$ the \emph{background} class, that is not an element of $\mathcal{Y}$.

In this general setting, we wish to create a model $f(x_i^k,\theta)\in \mathcal \Reals^{C+1}$ that predicts a posterior probability for each of the C classes and the background ${\hat{y}_{bg}}$, and we denote with $f(x_i^k,\theta)_y$ the posterior of class $y$. We wish to train such a model with all available databases $\{\mathcal{D}_k\}_{k=1}^{K}$.

In this setting, a label-contradiction problem arises due to inconsistent use of the background class for annotation protocols of each database: As is the common annotation practice, when annotating $\mathcal{D}_k$, all samples that in reality belong to one of the non-annotated label-set $\mathcal{Y}_{o,k}$ are assigned the background label $\hat{y}_{bg}$. As a result, the same sample may be given class $y_c \in \mathcal{Y}_k$ according to one annotation protocol, and $\hat{y}_{bg}$ according to other. Fig.~\ref{fig:intro} illustrates this. Consequently, standard learning frameworks, such as training a neural network $f$ with cross entropy (CE), will assign contradicting penalties to predictions about samples of the same content (e.g. same anatomy) when processing samples from different databases. We below describe the methods we studied to alleviate this.

\subsection{Adaptive cross entropy for learning from data with heterogeneous annotations}
\label{subsec:ace}

To train a model while avoiding the label-contradiction problem due to the differing definition of the `background' class, we need a learning framework that treats this class differently per database. Such a learning objective is \emph{adaptive cross entropy} (ACE) \cite{roulet2019joint}. It has been originally formulated for learning with two databases and the assumption that their label-sets $\mathcal{Y}_k$ are disjoint. We observe that ACE can be straightforwardly generalized to the case of learning from any number of databases with potentially overlapping label-sets:

\begin{equation} \label{eq:ace_total}
    J_{ace}(\{\mathcal{D}_k\}_{k=1}^K\}) = \sum_{k=1}^{K} \sum_{i} H_{ace}(x_i^k, y_i^k)
\end{equation}
\begin{equation} \label{eq:ace}
    H_{\textrm{ace}}(x_i^k,y_i^k) = 
    \begin{cases}
    -\log(f(x_i^k,\theta)_{y_i^k}) & \text{if } y_i^k \in \mathcal{Y}_{k} \\
    -\log(\sum_{y \in \{\hat{y}_{bg}\} \cap \mathcal{Y}_{o,k}} f(x_i^k, \theta)_{y}) & \text{otherwise} \\
    \end{cases}
\end{equation}

Here, $H_{ace}$ is entropy per sample and $J_{ace}$ the total cost across all databases. Intuitively, for every database $\mathcal{D}_k$, ACE behaves similar to CE for annotated samples. For samples that the annotation protocol of $\mathcal{D}_k$ leaves non-annotated (as background), it sums up the predicted probabilities for non-annotated classes $\mathcal{Y}_{o,k}$ and class $\hat{y}_{bg}$, forcing them to sum up to 1 (minimize $-log$). Fig.~\ref{fig:ace} illustrates this. How does this facilitate learning? It does not penalize a model for predicting any of the non-annotated classes $\mathcal{Y}_{o,k}$ for samples that have not been annotated and, therefore, not contradicting information learned from other databases where these classes are labeled. As a result, it enables making use of all available supervision signal from any sample annotated across databases.

We identify that for samples not annotated in $\mathcal{D}_k$ (i.e. considered background in $\mathcal{D}_k$), ACE does not explicitly encourage better predictions for one of the non-annotated classes $\mathcal{Y}_{o,k}$. In fact, the lower part of Eq.~\ref{eq:ace} will be minimized for any combination of posteriors that sum up to 1. We improve this by adopting ideas from semi-supervised learning and introducing them to ACE, as described next.

\subsection{Learning from non-annotated regions via Mean Teacher}
\label{subsec:mean_teacher}

% \comment{Components that help: a) EMA, b) augmentation. Refer to Fig.~\ref{fig:method3}.}

We here interpret samples that are assigned the background class in each database $\mathcal{D}_k$ as unlabeled samples, and investigate the integration of semi-supervised learning (SSL) in a framework for learning from heterogeneously labeled databases.

We study one of the most successful recent methods for SSL, the Mean Teacher (MT) \cite{tarvainen2017mean}. In a SSL setting, it assumes a labeled $\mathcal{D}_L$ and an unlabeled database $\mathcal{D}_U$.
It benefits from unlabeled data by learning model parameters $\theta$ such that predictions are consistent regardless perturbations of the input or the model parameters. This has been shown to improve generalization.

This is accomplished in MT via complementing a standard classification cost $J_{cl}$ (e.g. cross entropy $H_{ce}$) with a consistency cost $J_{con}$. The original definition of MT's cost function \cite{tarvainen2017mean} for SSL is given by the following: 

\begin{equation}
    J_{\textrm{mt}}(x_i, y_i) = J_{\textrm{cl}}(x_i, y_i) + J_{\textrm{con}}(x_i) 
\end{equation}
\begin{equation} \label{eq:mt_classif_loss}
    J_{cl}(x_i, y_i) = H_{ce}(x_i, y_i)  =
    \begin{cases}
    -\log(f_{stu}(x_i,\theta)_{y_i}) & \text{if } x_i \in \mathcal{D}_{L} \\
    0 & \text{if } x_i \in \mathcal{D}_{U}  \\
    \end{cases}
\end{equation}
\begin{equation}
    J_{con}(x_i ) = 
    (f_{stu}(x_i,\theta) - f_{tea}(x_i,\theta))^{2} 
    \quad  \quad \forall{x_i}
\end{equation}

The consistency cost $J_{con}$ is defined via two perturbations of the sample's $x_i$ embedding: the embeddings by the \emph{student} $f_{stu}$ and the \emph{teacher} $f_{tea}$. The perturbed embeddings are the result of two components. First, the student uses the current state of model parameters $\theta$, whereas the teacher uses an exponential moving average (EMA) of their values, $\theta_{ema}$. The assumption is that EMA over parameters improves predictions similar to an implicit ensemble, and hence it will enforce the student to predict better. Secondly, student and teacher embeddings are computed via different perturbations of the signal. In our settings, as commonly done, this is computed for different values of dropout masks between $f_{stu}$ and $f_{tea}$. Our ablation study (Sec.~\ref{sec:experiments}) will investigate the influence of both.

The above formulation cannot be straightforwardly applied for the general case of partially annotated databases, because CE would suffer from the label-contradiction problem for the background class (Sec.~\ref{subsec:baselines}). We extend the framework to this setting by combining it with ACE. This can be done by using ACE ($H_{ace}$, Eq.~\ref{eq:ace}) as the classification loss in Eq.~\ref{eq:mt_classif_loss}, instead of CE ($H_{ce}$). This combines benefits of learning from all samples $x_i^k$ that are annotated for each database $\mathcal{D}_k$, with the use of consistency loss $J_{con}(x_i^k)$ for all samples $x_i^k \in \{D_k\}_{k=1}^K$, which includes non-annotated (background) samples. We hypothesize that the latter will offer orthogonal benefits to those from ACE, improving predictions of non-annotated samples in each database. The following empirical investigations investigates this hypothesis.

\section{Experiments}
\label{sec:experiments}

\subsection{Data and model configuration}

\noindent\textbf{DB1:} This is an internal database consisting of 40 3D CT scans of the abdominal region of patients with cervical cancer.
The scans consist of between 183 and 331 axial slices with $512 \times 512$ pixel resolution. They were acquired with a full-bladder drinking protocol with patients in supine position. $20$ samples were randomly chosen for training, and the remaining $20$ used for testing. DB1 is considered fully annotated in our experiments, defining $\mathcal{Y}=\mathcal{Y}_1$ with $6$ labeled classes, which consist of: Bladder, Rectum, Uterus, Bones, left and right Femoral heads.

\noindent\textbf{DB2:} Partially annotated, public database consisting of 30 3D CT scans of the abdominal region of patients with cervical cancer from the Synapse benchmark \cite{cervixDatabase}. The scans consist of between 125 and 237 axial slices of $512 \times 512$ pixels. They were acquired with a full-bladder drinking protocol with most patients in prone and some in supine position. This database was used only for training. This is considered the partially annotated database with $3$ of the $6$ classes labeled. Therefore, $\mathcal{Y}_2$ here consists of: Bladder, Rectum and Uterus. The remaining Bones, left and right Femoral head classes are non-overlapping ($\mathcal{Y}_{o,2}$).

\noindent\textbf{Pre-processing:} All images were resampled to 2mm isotropic resolution followed by intensity capping ($-200 \text{ to} +200$) and normalisation ($\mu=0 \text{ and } \sigma =1$). Scans were reoriented to simulate supine patient position where necessary. 

\noindent\textbf{Main model:} We use a 3D CNN, DeepMedic, previously used for a variety of segmentation tasks with promising performance \cite{kamnitsas2017efficient}. We employ the `wide' model variant publicly available \url{https://github.com/deepmedic/deepmedic}, v0.8.4) and otherwise use the default  hyper-parameters and model architecture.

\noindent\textbf{Configuration of methods:} Hyper-parameters of the explored methods were set based on original works. Additional settings include the maximum weight of the consistency cost (set to $1.0$) and its ``warm-up'' period ($10$ training epochs starting from zero in the third epoch linearly increasing to the maximum weight). This was found to improve training convergence in preliminary experiments.

\subsection{Results}
\label{subsec:results}

\begin{table}[t]
	\begin{center}
	\caption{Dice \% from studied methods. (* significant difference vs SL.1, $p\!<\!0.05$)}
	%\centering
	%\small 
	%\setlength{\tabcolsep}{4pt}
	\begin{tabular}{@{}lccccccc@{}}
	    %\toprule
		\textbf{Class} &\textbf{SL1} &\textbf{SL12} &\textbf{ACE} &\textbf{PL} &\textbf{MT} &\textbf{ACE/PL} &\textbf{ACE/MT} \\
        \midrule % Or \hline, it does not break the vertical line
		Bladder     &$89.6$ &$90.7$   &$90.7$   &$90.6$   &$89.2$   &$91.0$   &$91.5$ \\
		Rectum      &$73.7$ &$76.1$   &$75.2$   &$77.3$   &$78.3$   &$76.7$   &$78.2$ \\
		Uterus      &$60.1$ &$67.4$   &$67.9$   &$67.9$   &$66.8$   &$68.4$   &$68.8$ \\
		\midrule
		Bones       &$87.9$ &$81.1$   &$88.3$   &$88.7$   &$89.2$   &$88.6$   &$88.2$ \\
		Fem.Head L  &$87.5$ &$85.1$   &$88.2$   &$88.3$   &$90.3$   &$88.9$   &$89.8$ \\
        Fem.Head R  &$87.6$ &$84.9$   &$88.1$   &$88.2$   &$88.9$   &$88.5$   &$88.8$ \\
        \midrule % Or \hline, it does not break the vertical line
		Overlapping &$74.5$ &$78.1^*$ &$77.9^*$ &$78.6^*$ &$78.1^*$ &$78.7$   &$79.5^*$ \\
		Non-Overlap.&$87.7$ &$83.7$   &$88.2^*$ &$88.4^*$ &$89.5^*$ &$88.7^*$ &$88.9^*$ \\
		Total Mean  &$81.1$ &$80.9$   &$83.1^*$ &$83.5^*$ &$83.8^*$ &$83.7^*$ &$84.2^*$ \\
		\bottomrule
	\end{tabular}
	\label{tab:dice_eval}
	\end{center}
\end{table}

% \comment{Describe the results. Provide tables and graphs. If the experiments have be broken down by "theme" (e.g. 2 different databases, or evaluating different parts of the method), then also break down this subsection. E.g. it can be 2 subsections (Results on DB1; Results on DB2; etc).}

% \comment{Results that compare the method to baselines and previous methods. Results shown in Table~\ref{tab:dice_eval}. Also, related results are also shown in Fig.~\ref{fig:evaluation}, which demonstrates that this and that happens.} 
All the below experiments were repeated for 3 seeds. We report average performance on DB1 test data (Dice\%) for all methods in Table~\ref{tab:dice_eval}.

\noindent\textbf{Baselines:}
We first evaluate a DeepMedic model trained only with supervised learning on fully labeled DB1 data. This \textbf{SL1} method performed well for segmenting abdominal tumours and organs, marking a suitable point for baseline comparison. The \textbf{SL12} method naively uses both databases for training a model with CE. Results for SL12 show clear improvements for overlapping classes over SL1. Performance for non-overlapping classes, however, is negatively affected. We hypothesise this is due to label contradiction across databases.

\noindent\textbf{Adaptive cross entropy:} We assess how well \textbf{ACE} \cite{roulet2019joint} mitigates the effect of label contradiction. We train DeepMedic with ACE using both DB1 and DB2. Accuracy for overlapping classes is maintained as with SL12, without losing accuracy for non-overlapping classes compared to SL1, confirming its effectiveness.

\noindent\textbf{Pseudo-labelling:} As additional comparison, we apply the pseudo-labelling SSL approach \cite{lee2013pseudo}. Here, predictions for DB2 from supervised SL1 are combined with partial annotations of DB2 to generate \emph{pseudo-labels} for DB2. This is done by over-writing the background class in the manual annotations for pixels where the model predicted a non-annotated class ($Y_{o,2}$). Then, a new model is trained using DB1 labels and DB2 pseudo-labels. This \textbf{PL} approach shows small improvements over ACE on average. We note that, contrary to ACE, it cannot be easily generalised to K databases as it requires K initial models and fusion of their predictions to form a single pseudo-label, which is not trivial. We also combine PL with ACE, simply by using predictions from the ACE method to create pseudo-labels. This \textbf{ACE/PL} approach improves over PL and ACE.

\noindent\textbf{Mean Teacher:} We first evaluate the \textbf{MT} approach in a purely semi-supervised fashion. We train MT via Eq.~\ref{eq:mt_classif_loss}, using CE on DB1 as labeled $\mathcal{D}_L$, and DB2 as \emph{completely unlabeled} $\mathcal{D}_U$ via the consistency loss only. In addition to EMA as a signal perturbation, $50\%$ dropout is used in all layers except the first 2. MT shows clear improvements over SL1, and modest improvements over ACE and PL, even though it does not use \emph{any} labels from DB2, contrary to ACE and PL.

\noindent\textbf{Combined ACE and Mean Teacher:} Finally, we evaluate the proposed combination of ACE with MT, \textbf{ACE/MT}, taking advantage of their complementary nature. We use the DB2 partial labels directly within the MT framework through the ACE loss, instead of CE. The results of ACE/MT show best overall performance across all studied methods. Overall, ACE/MT improves over SL1 baseline by $3\%$ DSC, and over the most recent approach for this problem, ACE, by $1\%$ DSC, supporting that SSL provides complementary benefits.

\begin{figure}[t]
	\centering
% 	\captionsetup{font=scriptsize}
\begin{subfigure}[b]{0.49\textwidth}
	\centering
	\tiny
	\begin{tabular}{@{}lcc|ccc@{}}

	    %\toprule
		\textbf{Class} & \textbf{ACE} & \textbf{ACE$+$} & \textbf{MT$_t$} & \textbf{MT$_s$} & \textbf{MT$_s+$} \\
        \midrule % Or \hline, it does not break the vertical line
		Bladder     &$90.7$ &$90.6$ & $89.6$ &$89.6$ &$89.2$ \\ 
		Rectum      &$75.2$ &$76.2$ & $77.0$ &$76.8$ &$78.3$ \\
		Uterus      &$67.9$ &$67.5$ & $63.7$ &$63.7$ &$66.8$ \\
		\midrule
		Bones       &$88.3$ &$88.3$ & $88.3$ &$88.3$ &$89.2$ \\
		Fem.Head L  &$88.2$ &$88.2$ & $88.8$ &$88.9$ &$90.3$ \\
        Fem.Head R  &$88.1$ &$87.3$ & $89.1$ &$89.0$ &$88.9$ \\
        \midrule % Or \hline, it does not break the vertical line
		Overlapping &$77.9$ &$78.1$ & $76.8$ &$76.7$ &$78.1$ \\
		Non-Overlap.&$88.2$ &$87.9$ & $88.7$ &$88.7$ &$89.5$ \\
		Total Mean  &$83.1$ &$83.0$ & $82.8$ &$82.7$ &$83.8$ \\
		\bottomrule
	\end{tabular}
	\caption{Results (DSC\%) of ablation study}
	\label{table:ablation}
\end{subfigure}
\begin{subfigure}[b]{0.33\textwidth}
	\centering
	\includegraphics[clip=true, trim=0pt 0pt 0pt 0pt,  width=1.0\textwidth]{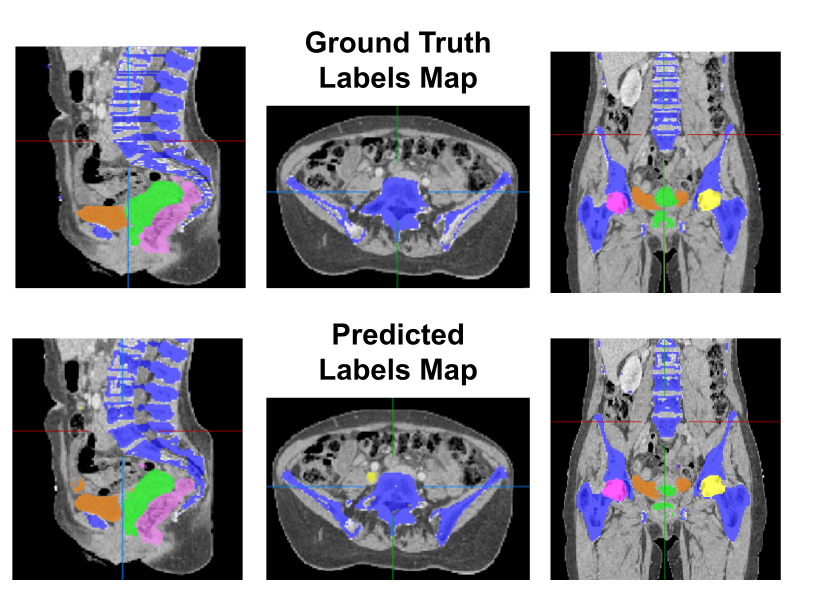}
	\caption{ACE/MT predictions}
	\label{fig:results3}
\end{subfigure}
\caption{(a) Ablation study on mean teacher. We find that predictions via (EMA) teacher (MT$_t$) and student(MT$_s$) parameters are similar. Training with more dropout (MT$_s+$) improves the framework. The baseline ACE does not improve by increased dropout (ACE$+$). Therefore we conclude it is the interplay of consistency loss and perturbations that MT benefits from. (b) Example of results obtained by best method, combination of ACE and Mean Teacher.
}
\label{fig:evaluation}
\end{figure}

\noindent\textbf{Ablation study:} 
We perform an ablation study on MT to test whether benefits are provided due to EMA or perturbation via dropout. Results (Dice\%) are summarized in Figure~\ref{table:ablation}. Specifically, we first train a model with MT using dropout 50\% only on the 2 last hidden layers. We report performance of predictions made using the student parameters (\textbf{MT}$_s$) and the (EMA) teacher parameters (\textbf{MT}$_t$). We also trained MT with more perturbation, using dropout $50\%$ on all layers except first 2 (\textbf{MT}$+$). We find that the EMA parameters make no difference. In contrast, additional dropout offers substantial improvements. We test whether the additional dropout benefits ACE, and find no improvements (\textbf{ACE} vs \textbf{ACE}$+$). Therefore, we conclude it is the interplay of MT's consistency loss with perturbation that leads to the method's high performance.

\section{Conclusion}

% \comment{One paragraph that summarizes what we did, what we have shown, and why it is important. Usually approx 10 lines, as there is not much more space in a MICCAI paper. THIS MUST END AT 8 PAGES. Not a single line in page 9.

% The REFERENCES can go up to end of page 10. But the main paper (everything except references) must end at end of page 8.}

This study investigated several strategies for learning from databases that were annotated via different annotation protocols, resulting in partial overlapping sets of labels. In the process, we identified that a semi-supervised learning approach, Mean Teacher \cite{tarvainen2017mean}, offers complementary benefits with a recently proposed approach for the task, adaptive cross entropy \cite{roulet2019joint}. 
We demonstrated that their combination is elegant and effective, outperforming its individual components.
Experiments on the task of segmenting anatomical structures in abdominal CT for cervical cancer radiotherapy planning demonstrated that this proposed combined approach can successfully leverage an internal and a public database with partial overlap of labels. It achieved a $+3\%$ Dice score improvement over a supervised model trained using only the internal database, which was specifically made for radiotherapy planning. Our results demonstrate the potential of these methods, which enable leveraging public, heterogeneously annotated datasets in order to overcome the scarcity of high quality annotated data.

% We verified the benefit of augmentation for the MT approach however we only experimented with differing levels of dropout. Various other augmentation approaches could be trialled and evaluated to see which best improved performance for the approach. Virtual Adversarial Training in another compelling approach that is compatible with the mean teacher approach that chooses the most adversarial augmentation to apply to the model. In this way the augmentation does not need to be decided a priori instead chosen at each stage of training to be the most adversarial. 

%\comment{Label annotation shift is not the only difference between databases that may cause problems for combined learning. Other inconsistencies worth investigating include: (1) differing acquisition approaches (2) varied target tasks (3) unrelated patient cohort groups. Beyond these, a more in depth understanding of the improvements shown, through evaluation in a real world setting, would be insightful.} 

\section*{Acknowledgements}

This work received funding from the UKRI London Medical Imaging \& Artificial Intelligence Centre for Value Based Healthcare and the European Research Council (ERC) under the European Union's Horizon 2020 research and innovation programme (grant agreement No 757173, project MIRA, ERC-2017-STG). AT receives a grant from Lady Garden Foundation.

\bibliographystyle{splncs03}
\bibliography{references}

\end{document}